# Path of Vowel Raising in Chengdu Dialect of Mandarin[1]

Hai Hu, Yiwen Zhang
*Indiana University Bloomington*

He and Rao (2013) reported a raising phenomenon of /a/ in /Xan/ (X being a consonant or a vowel) in Chengdu dialect of Mandarin, i.e. /a/ is realized as [ɛ] for young speakers but [æ] for older speakers, but they offered no acoustic analysis. We designed an acoustic study that examined the realization of /Xan/ in speakers of different age (old vs. young) and gender (male vs. female) groups, where X represents three conditions: 1) *unaspirated consonants* C ([p], [t], [k]), 2) *aspirated consonants* $C^h$ ([$p^h$], [$t^h$], [$k^h$]), and 3) *high vowels* V ([i], [y], [u]). 17 native speakers were asked to read /Xan/ characters and the F1 values are extracted for comparison. Our results confirmed the raising effect in He and Rao (2013), i.e., young speakers realize /a/ as [ɛ] in /an/, whereas older speakers in the most part realize it as [æ]. Also, female speakers raise more than male speakers within the same age group. Interestingly, within the /Van/ condition, older speakers do raise /a/ in /ian/ and /yan/. We interpret this as /a/ first assimilates to its preceding front high vowels /i/ and /y/ for older speakers, which then becomes phonologized in younger speakers in all conditions, including /$C^h$an/ and /Can/. This shows a possible trajectory of the ongoing sound change in the Chengdu dialect.

## 0 Introduction

He and Rao (2013) report a raising phenomenon in Chengdu dialect of Mandarin. Specifically, native speakers born in the 1950s pronounce the phoneme /a/ as [æ] in nasal environment /an/, whereas younger generations (born after the 1980s, roughly) raise [æ] to [ɛ] in most instances. The degree of raising from [æ] to [ɛ] seems to be also related to the environment for /an/. It is more likely to occur in environments where there are adjacent high vowels [i], [u] or [y]. He and Rao (2013) also report that the raising effect originates from conditions where the consonant preceding /an/ is aspirated.

This acoustic study tries to examine such raising effect by comparing the vowel height of /an/ in different phonological environments of two age groups, the young and old. We will also look at the role of gender, and phonological environment in the raising phenomenon.

[1] We thank Prof. Ken de Jong and Phillip Weirich for their help throughout the project. The first author is partially supported by China Scholarship Council.

The results show that in general the raising of /an/ in young speakers is confirmed, but the aspirated-unaspirated contrast is not born out. We also discuss the implications of the study and briefly discuss the possible path for raising in the Chengdu dialect.

## 1 Literature review

In this section, we review previous literature on vowels in Chengdu dialect with a focus on the rhyme /an/, and how others have measured vowel raising phenomena in other languages.

| Year of Analysis | 1941 | 1956 | 1956 | c.a. 1982 | c.a. 1983 | c.a. 2006 |
|---|---|---|---|---|---|---|
| Published in | Yang (1984) | Zhen (1958) | Zhen, Hao and Chen (1960) | Liang (1982) | Zhen (1983) | He and Rao (2013) |
| /iai/ | a | ɛ | ɛ | ɛ | NA | ɛ |
| /ian/ | e | æ | ɛ | ɛ | ɛ̃ | æ / ɛ |
| /yan/ | e | æ | ɛ | ɛ | ɛ̃ | æ / ɛ |
| /Can/ | a | A | æ | NA | ã | æ / ɛ |
| /uan/ | a | A | æ | NA | ã | NA |

Table 1: /an/ in previous studies

### 1.1 Literature on vowels in Chengdu dialect

Chengdu is the capital city of Sichuan Province in southwest China; Chengdu dialect is usually categorized as Southwest Mandarin, which is similar to Standard Mandarin. In Chengdu dialect, one syllable usually corresponds to one morpheme, allowing us to use mono-syllabic tokens in the recording. There are 6 possible syllable patterns: CV, CVN, CVV, CVVN, V and VN. /an/ can appear in CVN, CVVN and VN.

Many previous studies have documented the realization of /an/ (see Table 1). From Table 1, we see almost a three-way distinction:

/iai/ – /ian/ /yan/ – /Can/ /uan/

It can be drawn that historically /iai/ is almost always realized as [ɛ]. On the other hand, /ian/ and /yan/ are gradually raised whereas /Can/ and /uan/ seem to be raised only very recently, if raised at all. He and Rao (2013) is the first and possibly the only literature that documents the raising of /Can/, despite the fact that such raising effect has been noticed by many new comers to Chengdu who immediately notice that Chengdu people pronounce /pan/ or /fan/ very differently from people in nearby cities. He and Rao (2013) report that female speakers born after 1980s exhibit a strong raising of /a/ in /an/, and male speakers born after 1990s generally raise /a/ in /an/. Thus female speakers seem to lead the vowel change.

In addition, He and Rao (2013) also reports that the raising is more prominent when /an/ is preceded by an aspirated consonant, but no reason is provided. We will test this in our study by comparing the vowel height of /Can/ and /C$^h$an/.

However, they do not mention if their study is an acoustic analysis. Apart from that, very few studies have investigated the age difference and phonological environment of the Chengdu vowel raising. Therefore, we feel the need to document this possible vowel change.

## 1.2 Vowel raising measurement

### 1.2.1 /æ/ raising in GA Northern Cities Chain Shift

Many studies have looked at vowel raising in different languages/varieties. Clopper, Pisoni and de Jong (2005) collected data from 6 dialectal regions of the US and plotted vowel charts of these 6 regions. From their vowel chart, we can clearly see the raised /æ/ in Northerners which is part of the Northern Cities Chain Shift. Statistical analysis such as post-hoc Tukey is also used to confirm the raising of /æ/. In addition, they used Labonov normalization for all speakers (Labonov, 1971).

### 1.2.2 New Zealand vowel raising

Watson, Maclagan and Harrington (2000) compare recordings of 1948 to 1995 to see if the vowels of New Zealand English (NZE) have changed, particularly whether /ɛ/ has been raised or not.

Two methods are used to determine the raising of /ɛ/. First, they use /i/ and /æ/ as reference vowels and discover that in Old NZE /ɛ/ is almost in the mid point of /i/ and /æ/, whereas in Modern NZE, /ɛ/ is very close to /i/, which indicates the raising of /ɛ/. Second, they use t-test on the F1 and F2 values of the vowels they are investigating. The significance difference found between /ɛ/ in Old and Modern NZE confirms the raising. Labonov normalization is also used in their study.

In sum, two methods are commonly used in determining vowel raising: 1) plotting vowel chart and eyeballing, 2) statistical test, be it t-test or post-hoc Tukey tests in ANOVA.

# 2 Method

## 2.1 Research questions

Based on previous literature, we ask the following research questions in this study.

1. **Age**: Does the young age group raise /a/ more than the old age group when /a/ is followed by a nasal coda /n/? That is, are there differences in the height of /a/ in the experimental conditions of /an/ between the young and older age groups?
2. **Gender**: Within each age group, do female subjects exhibit more raising?
3. **Phonological environment**: Is the height of /a/ different in different phonological environments?

## 2.2 Subjects

Altogether 21 native speakers of Chengdu dialect were recruited originally, of which three are older female speakers, four are older male speakers, eight are young female speakers and six are young male speakers. Four female subjects are excluded for poor quality of recording.

Background questionnaires are collected from the young subjects (unfortunately we were unable to collect background questionnaires from the old age group, but it is ensured that they are all native speakers of Chengdu dialect). All of them are native speakers of Chengdu dialect. The parents of all young female subjects are also born in Chengdu, whereas only two of the young male subjects' parents are born in Chengdu. The daily communication in homes of all subjects are Chengdu dialect. All young subjects have spent most of their lives (mostly more than 18 years except for ym4) in Chengdu (see Table 2).

| | No. | Age | Gender | ParentsChengdu? | YrsInChengdu |
|---|---|---|---|---|---|
| Young Age Group | ym1 | 27 | M | n | 19 |
| | ym2 | 27 | M | n | 18 |
| | ym3 | 27 | M | n | 20 |
| | ym4 | 25 | M | n | 6 |
| | ym5 | 27 | M | y | 22 |
| | ym6 | 27 | M | n | 18 |
| | yf1 | 26 | F | y | 18 |
| | yf2 | 26 | F | y | 18 |
| | yf3 | 26 | F | y | 18 |
| | yf4 | 28 | F | y | 22 |
| Old Age Group | om1 | 59 | M | NA | 59 |
| | om2 | 57 | M | NA | 38 |
| | om3 | 57 | M | NA | 57 |
| | om4 | 56 | M | NA | 56 |
| | of1 | 49 | F | NA | 49 |
| | of2 | 49 | F | NA | 49 |
| | of3 | 43 | F | NA | 43 |

Table 2: Language background of subjects

## 2.3 Material

### 2.3.1 Reference Vowels

From the literature review (see Table 1), /iai/ seems to be a good choice for reference vowel. However, two reasons exclude /iai/ as the possible reference vowel in this study. First, /iai/ can only appear after two consonants /ɕ/ and /t͡ɕ/ in the dialect, whereas the other four rhymes have a much more varied environment, thus making the comparison very limited. Second, to our knowledge, young speakers of the dialect seldom pronounce the rhyme /iai/ now, because of the influence of standard Mandarin. /ɕiai/ is now more often pronounced as /ɕiɛ/ by young speakers, most likely to be influenced by its pronunciation in standard Mandarin (see Zhou, 2001).

For these reasons, we choose /i/ and /a/ as the reference vowel. [ta] will be the lowest vowel in the vowel space, whereas [ti] will mark the highest vowel height. We will then compare the height of /a/ in /an/ with the two reference vowels to see if it is raised in the experimental conditions. Details will be introduced in section 2.4.

### 2.3.2 Experimental tokens

In order to see if there are differences among phonological environments, we designed three environments. The first is /Can/ and /C$^h$an/, which include both unaspirated and aspirated consonants, where /an/ is preceded by a consonant. This contrast is to test what He and Rao (2013) has reported. Namely the /an/ raising is more prominent when preceded by an aspirated consonant. The second is /Van/ which we termed "Diphthong". This is to test whether there is a difference in /an/ realization when it is preceded by a vowel, which corresponds to /ian/, /yan/ and /uan/ in Table 1. The last environment is the reference vowels. Apart from the highest point in the vowel chart /i/ and the lowest /a/, we also include /e/, /o/ and /u/ in order to plot the vowel chart of each participant.

The experimental tokens are presented in Table 3. We use three different characters for each syllable, indicated by the number after the syllable.

| Condition | Environment | | | | | | |
|---|---|---|---|---|---|---|---|
| | | | an | an | an | | |
| | Unaspirated | p | pan1 | pan2 | pan3 | | |
| | | t | tan1 | tan2 | tan3 | | |
| | | k | kan1 | kan2 | kan3 | | |
| | Aspirated | $p^h$ | $p^h$an1 | $p^h$an2 | $p^h$an3 | | |
| | | $t^h$ | $t^h$an1 | $t^h$an2 | $t^h$an3 | | |
| | | $k^h$ | $k^h$an1 | $k^h$an2 | $k^h$an3 | | |
| *EXPERIMENTAL* | | | | | | | |
| | | | ian | uan | yan | | |
| | | 0 | ian1 | uan1 | yan1 | | |
| | Diphthong | 0 | ian2 | uan2 | yan2 | | |
| | | 0 | ian3 | uan3 | yan3 | | |
| | | | a | i | u | e | o |
| *REFERENCE* | Reference | t | ta1 | ti1 | tu1 | te1 | to1 |
| | | t | ta2 | ti2 | tu2 | te2 | to2 |

Table 3: Experimental tokens in IPA

Altogether 37 experimental tokens are used in this study. In addition, in order to avoid interference between the experimental tokens, we added fillers ending with mid vowels /o/ or /oŋ/ between experiment tokens, since they are mid vowels which we expect will not interfere with the height of the other vowels we are examining. Altogether 70 tokens are recorded by each subject.

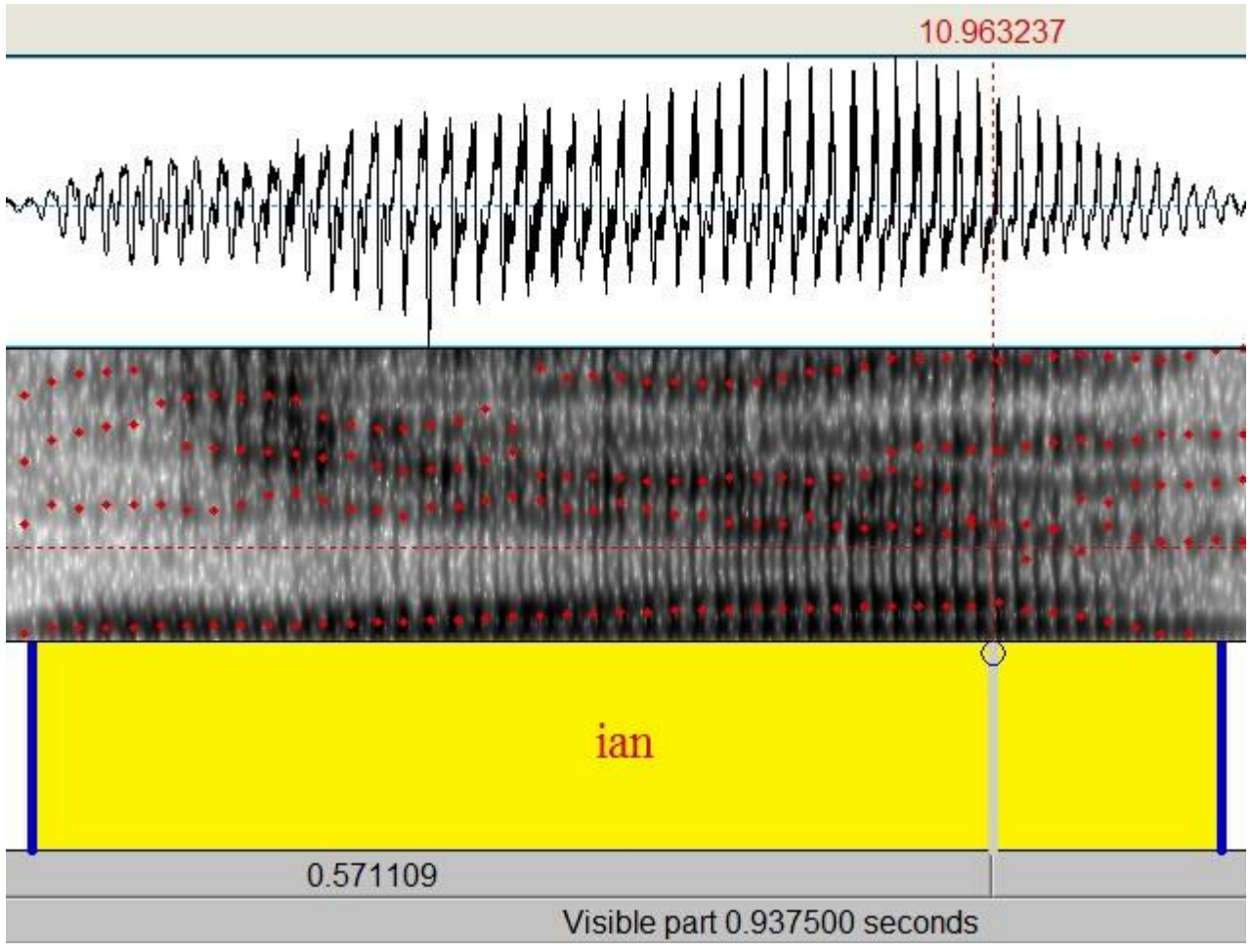

Figure 1: An example of annotation for /ian/.

## 2.4 Procedures

Now we describe the recording and analysis procedures.

### 2.4.1 Recording

Recordings are all done by the subjects reading a list of characters (in Chinese one character corresponds to one syllable and usually one morpheme). Recording apps on smart phones are used; these apps are capable of recording .wav format. All 17 recordings used in this study are in .wav format (i.e. mp3 recordings were deleted).

### 2.4.2 Labeling

Next, we use Praat (Boersma et al., 2002) to manually label the data. For "Unaspirated", "Aspirated" and "Reference" conditions the most stable part of the F1 is labeled. For the "Diphthong" condition (/ian/, /yan/ and /uan/), we labeled the highest point of F1 in the syllable. The reason is that /a/ is the lowest vowel in "Diphthong" condition (compared to /i/, /y/ and /u/) and should therefore have the highest F1 (see Figure 1).

Then we use a script to extract the F1 and F2 of the labeled points in all recordings.

### 2.4.3 Analyzing

Vowel charts are plotted using R (R Core Team, 2014).[1] They provide a direct visualization of the vowel spaces of the subjects, through which we can eyeball whether there is raising.

In addition, post-hoc Tukey tests on F1 are performed to determine whether there exists statistically significant differences between different age groups, genders and phonological environments. Then we will use the *diff* value to calculate the 'Height' of /an/ with reference to two reference vowels /i/ and /a/. That is, the F1 value of /i/ and /a/ will be the lower and upper bound of the F1 range; the F1 of /an/ of different groups will be plotted on to the scale and then normalized. For example, on a F1 scale of 250 Hz (/i/) - 900 Hz (/a/), if the F1 of /an/ is 700 Hz, the 'Height' of /an/ will be:

$$Height_{/an/} = \frac{F1_{/a/} - F1_{/an/}}{F1_{/a/} - F1_{/i/}} \times 100\% = \frac{900 - 750}{900 - 250} \times 100\% = 23.1\% \qquad (\star)$$

---

[1] We would like to thank Phillip Weirich for the help with R code.

Thus we can compare the different 'Height' of /an/ in different groups.

## 2.5 Hypothesis

Based on the research questions and literature review, we have the following hypotheses.

H1: Degree of raising is more significant for young group than for old group:

- The realization of /a/ is lower for old group than for young group.
- The degree of raising of /a/ from reference vowel to [an], [uan], [ian], [yan] is larger for young group than for old group.

H2: Raising is more significant for female group than for male group:

- Within each age group, the degree of raising is larger in female speakers.
- Within each gender, young speakers still raise /an/ more than the old speakers.
- The difference in the degree of raising between young and old female participants is larger than that between two age groups of male speakers.

H3: Raising is influenced by phonological environment:

- The height of /a/ is higher in environments where there is an immediately preceding high vowel (i.e. Diphthong [uan], [ian] and [yan]).
- The height of /a/ is higher in aspirated environments (i.e. [$p^h$], [$t^h$], [$k^h$]) than in unaspirated environments (i.e. [p], [t], [k]).
- The height of /a/ is different among [uan], [ian] and [yan]. Specifically, /a/ in [ian] and [yan] is higher than /a/ in [uan].

# 3 Results

In this section, we report both the statistical results and the descriptive analysis of our data.

## 3.1 Statistical results

### 3.1.1 Between age groups

First we test whether it is true that the young age group exhibit more raising in all experimental environments. That is, "Unaspirated", "Aspirated" and "Diphthong" will together be collapsed under the *experimental condition* (i.e. "_an"), and it will be compared to the *reference vowel* /a/ (see Figure 2).

A two way ANOVA with condition (experimental "_an", reference "a") and age (young, old) reveals a main effect of condition on the value of F1, F = 478.98, p<.001, and a main effect of age on the value of F1, F = 73.18, p<.001. It also reveals an interaction between condition and age, F = 4.48, p=.01. Post-hoc analysis using Tukey's HSD indicates that:

- Between young and old group, the value of F1 of [a] in experimental condition for young age group is statistically lower than [a] of old group (diff =-101.31, p <.001). This indicates that the vowel height of /an/ in the younger age group is higher than that of the old age group, thus suggesting the young speakers are raising /an/ compared to their parents' generation, which is captured by the red box for condition "_an" condition in the middle of Figure 2.
- Within both the young and old age group, the value of F1 of [a] in experimental condition is significantly lower than the F1 of reference vowel [a] (diff =-104.56, p =.03). This suggests that in both age groups the height of /an/ is significantly higher than reference vowel /a/.
- Finally, there is no significant difference in the F1 value of Reference vowel [a] between two age groups (p =.98), thus indicating that the young and old speakers have a similar vowel height in the reference condition. This is shown in the similar length of the green boxes of both age groups in Figure 2.

Using the equation (⋆) and the mean value from Table 4, we can calculate the mean 'Height' of /an/ in both the young and old age group. The result is that the $\text{Height}_{/an/}$ for young age group is

$$\frac{926.2534 - 706.0808}{926.2543 - 302.7332} \times 100\% = 35.3\%$$

whereas the $\text{Height}_{/an/}$ for old age group is

$$\frac{911.9501 - 807.3867}{911.9501 - 344.5206} \times 100\% = 18.4\%$$

This shows that the experimental /an/ in young age group is at about a third (35.3%) in height on the scale established by their two reference vowels /i/ and /a/. However, the experimental /an/ in old age group is at a position of 18.4% on the reference scale between /i/ and /a/. This lends further support for our argument that the younger age group raises /an/ more than the old age group.

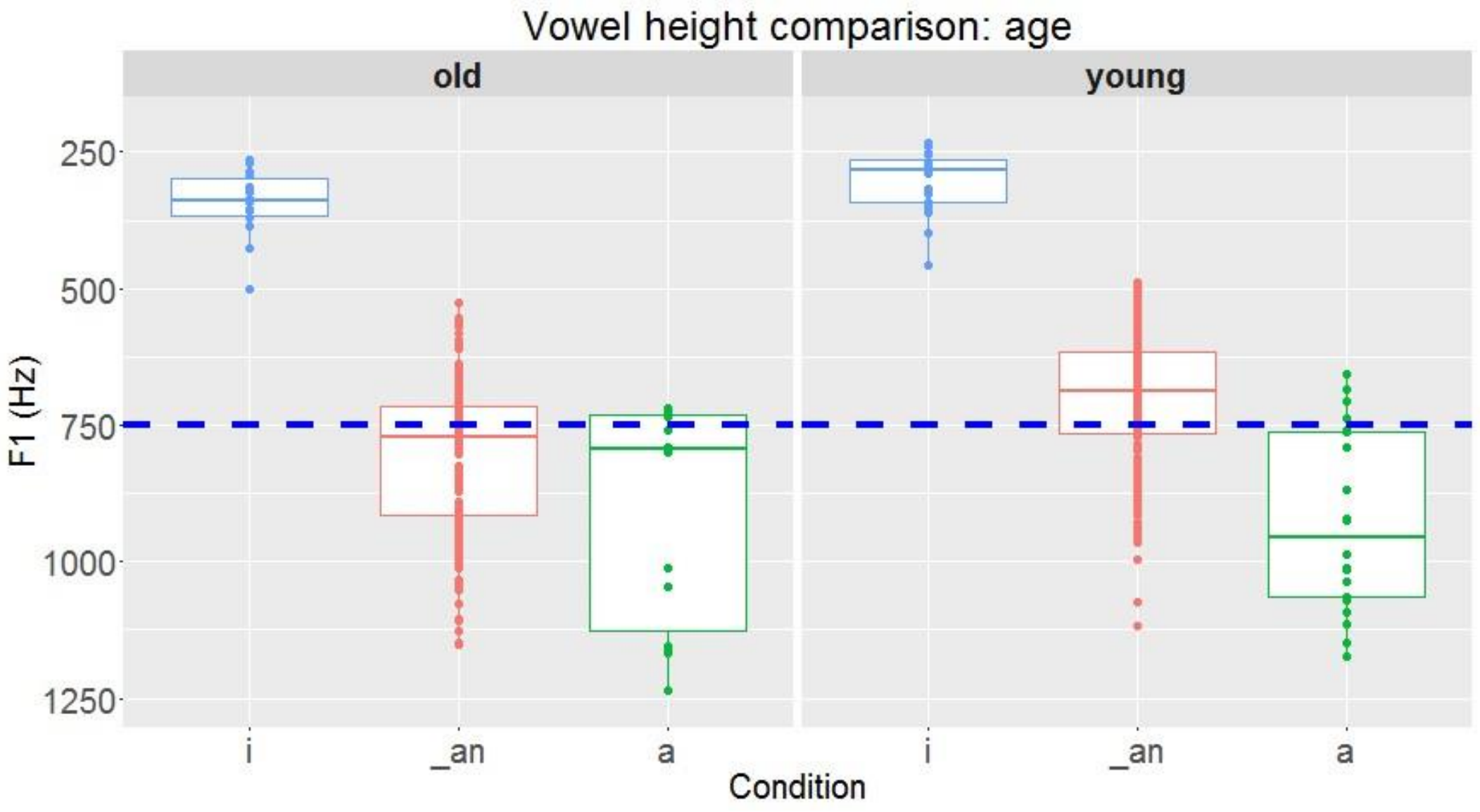

Figure 2: Old vs. young age group.

| Condition | Age group | F1(Hz) |
|---|---|---|
| an | old | 807.3867 |
| an | young | 706.0808 |
| a | old | 911.9501 |
| a | young | 926.2534 |
| i | old | 344.5206 |
| i | young | 302.7332 |

Table 4: Mean F1 of Experimental /an/, Reference /a/ and /i/

### 3.1.2 Between two genders

A three way ANOVA with condition (Experimental /_an/, Reference /a/) age (young, old) and gender (male, female) reveals a main effect of condition, age and a main effect of gender on the value of F1 of [a], F = 414.27, p<.001. It also reveals an interaction between condition and gender, $F = 15.99$, $p<.001$; and interaction between age and gender, $F = 24.19$, $p<.001$; but no interaction between condition, age and gender, $F = 0.67$, $p=.57$. These are shown in Figure 3. Post-hoc analysis using Tukey's HSD indicates that:

- Within each age group, the female speakers raise /an/ more than the male speakers. That is, in the **young** age group, the difference in F1 value of Experimental /an/ and Reference /a/ for female speakers is 295.00 (p=.000); the difference for the male speakers is 75.97 (p=.001). This shows clearly that the female speakers in young age group raise much more than the male speakers.

In the **older** age group, the difference in F1 value of Experimental /an/ and Reference /a/ for female speakers is 206.34 (p=.000); the difference for the male speakers is 28.23 (p=.999). This shows that the female speakers also raise much more than male speakers in the old age group.

- If we compare within each gender group, the young speakers still raise more than the old speakers. That is, within the female group, the degree of raising is 206.34 for old speakers but 295.00 for young speakers. Within the male group, the degree of raising is 28.34 for old speakers but 75.97 for young speakers.

### 3.1.3 Between phonological environments

A three-way ANOVA with gender (male, female), age (young, old) and environment (Diphthong, Aspirated, Unaspirated, Reference /a/) reveals a main effect of gender, a main effect of age and a main effect of environment on the value of F1 of [a], $F$ = 365.22, $p$<.001. It also reveals an interaction between age and environment, $F$ = 3.802, $p$ =.002; an interaction effect between gender and environment, $F$ =10.51, $p$ <.001; but no interaction effect among age, gender and environment, $F$ =.49, $p$ =.78. Post-hoc analysis using Tukey's HSD indicates that: Within each gender and age group, the value of F1 of [a] in Diphthong, Aspirated and Unaspirated are not statistically different from each other ($p$ >.05). These are also demonstrated in Figure 3.

Crucially, the fact that no significant difference between the aspirated and the unaspirated environments (the yellow and pink bars in Figure 3) is discovered does not support the analysis in He and Rao (2013) where aspirated environments are more prone to raising.

A one way ANOVA with high vowel-preceding environments (i.e. [uan], [yan], [ian]) and a post-hoc analysis using Tukey's HSD indicates that the F1 value of /a/ in [yan] is statistically significantly lower than [uan] ($F$ =-92.28, $p$ <.02). There is no statistically significantly difference between the F1 value of /a/ in [uan] and [ian] as well as in [yan] and [ian]. This is shown in Figure 4, and provides partial support for previous literature (see Table 1) to group [yan] and [ian] together, but exclude [uan].

## 3.2 Descriptive summary

Two representative vowel charts are shown in Figure 5. We see that in the representative old male speaker the reference /a/ is of the same height to experimental tokens of different phonological conditions. Sometimes it is even higher (e.g. compared with /uan/). However, the representative young female speaker demonstrates neatly the raising phenomenon, as her /_an/ is much closer to /e/ than to /a/. The remaining question is that whether it is possible that the old male speaker has raised /a/.

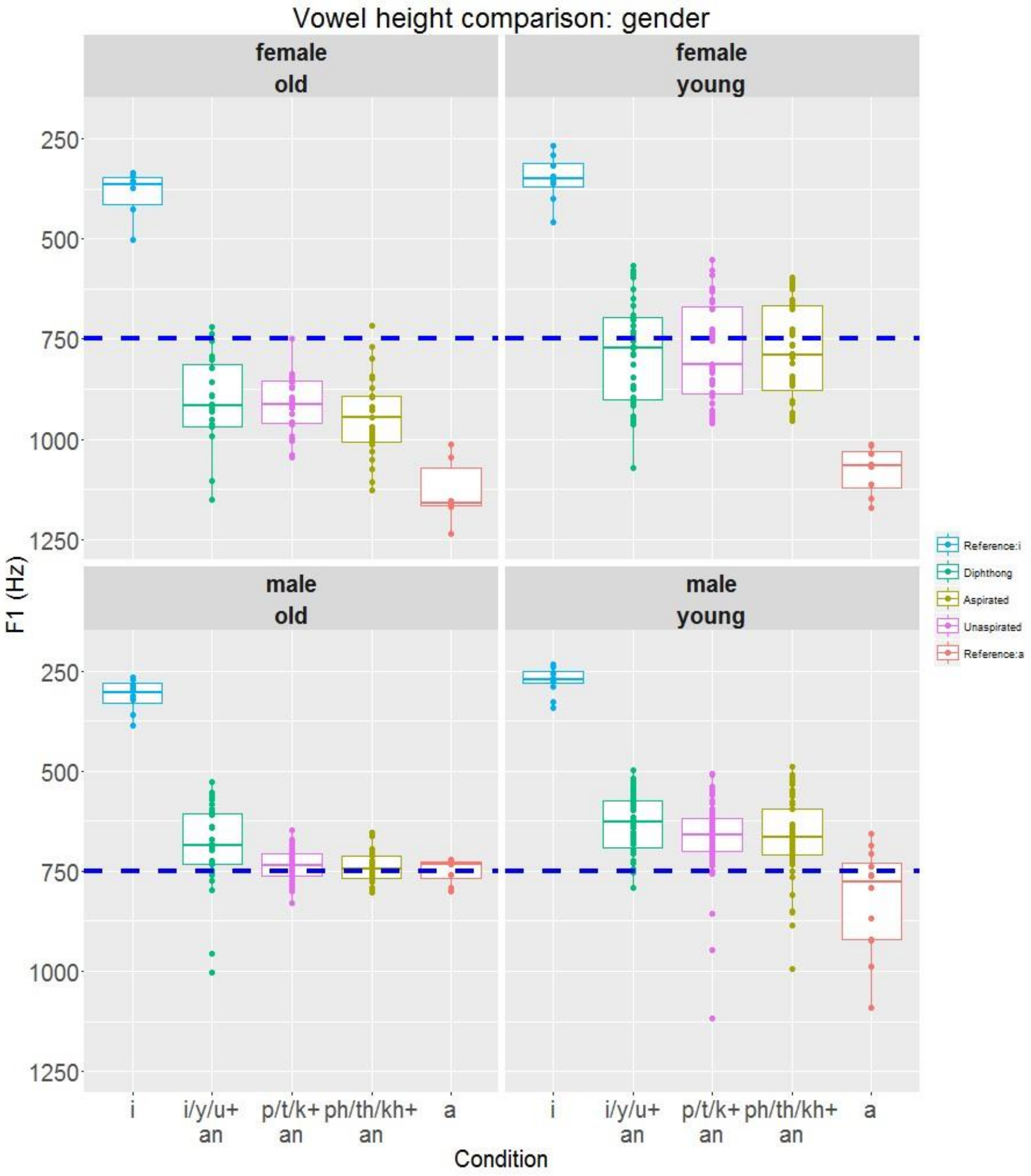

Figure 3: Comparison across age, gender and phonological environment.

When we look at Figure 6 where the vowel chart of the mean of two age groups are plotted one on top of another, it is more straight-forward that the young age group (green) raises the experimental /_an/ whereas the old age group (red) does not. The two reference vowels /i/ and /a/ are almost in the same position on the chart, suggesting that they are stable reference points for the study.

Another very intriguing point in Figure 6 is that for the old age group, their /ian/ and /yan/ are almost as high as green cluster of /an/ for young speakers, whereas their /uan/ has the lowest height. This seems to suggest that /ian/ and /yan/ are the first to raise and /uan/ is the last one to join the raising phenomenon. What Figure 6 shows is particularly interesting as it demonstrates the possible order of raising for different conditions.

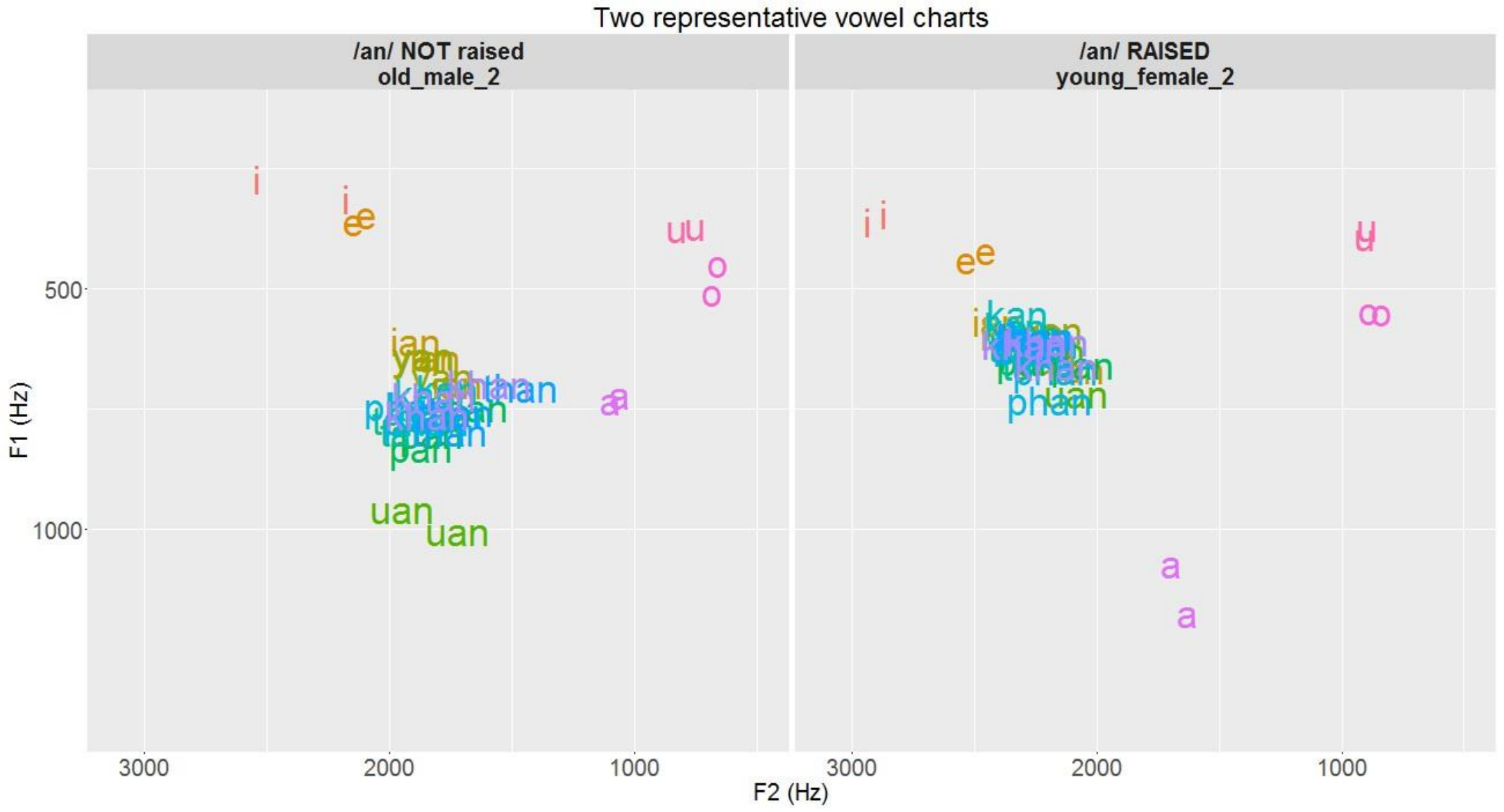

Figure 5: Two representative vowel chart.

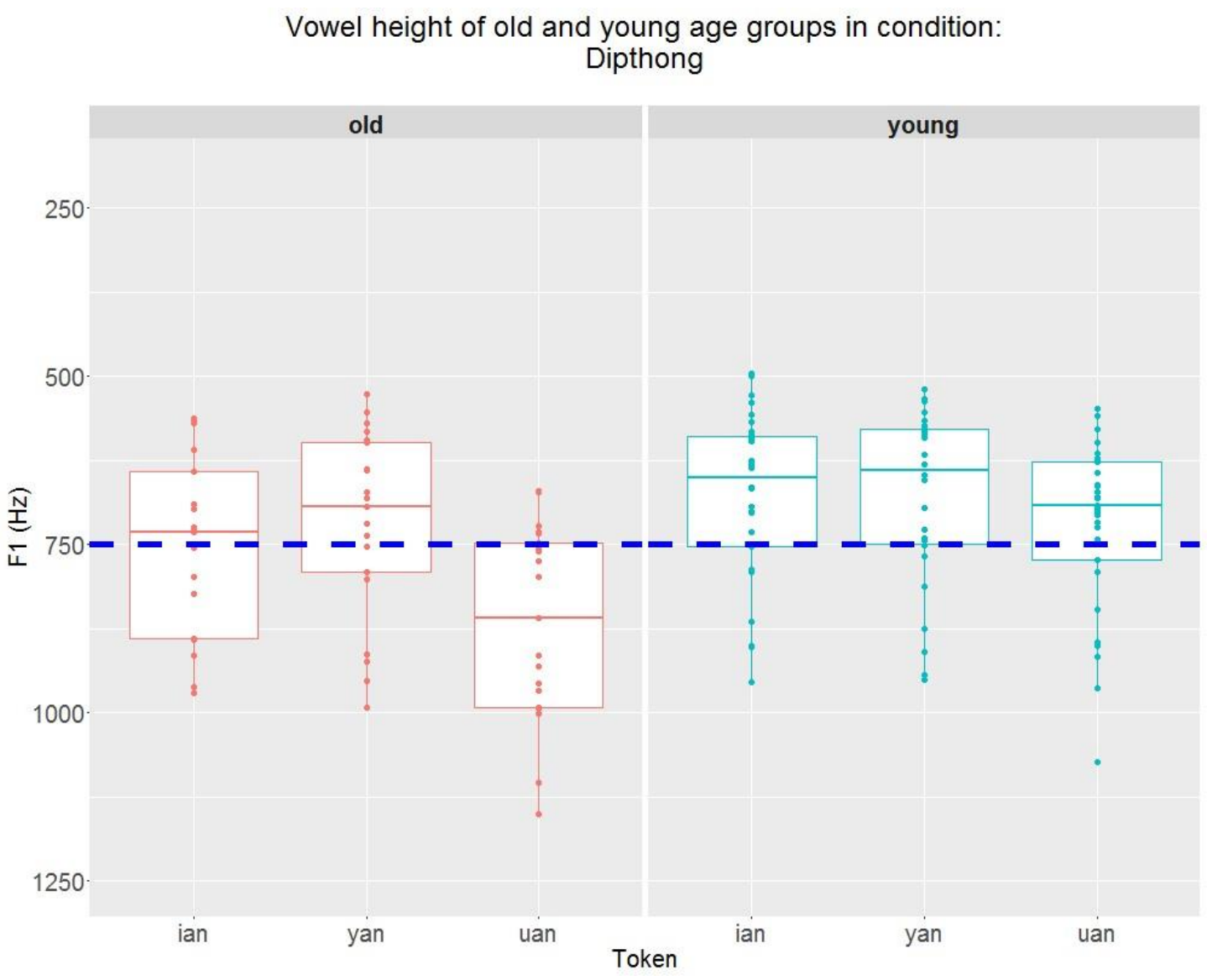

Figure 4: Comparison between old and young age groups in i/y/u+an.

To sum up, our data shows that:

1. Raising is more significant for the young age group than the old group, supporting our first hypothesis H1. Specifically:
    - The height of the vowel in experimental condition, i.e. /_an/ in the young age group is higher than that of the old age group.

- The degree of raising of from the reference vowel /a/ to experimental condition /_an/ is greater for the young age group.

2. Raising is more significant for the female group than for the male group within each age group, supporting H2. Specifically:

   - The degree of raising of /a/ from reference condition to experimental condition is larger for female participants than for male participants in both young and old age groups.
   - The contrast between young and old age groups is still attested within each gender group. That is, the young female speakers still have more raising than the old female speakers. The same is true for male speakers.

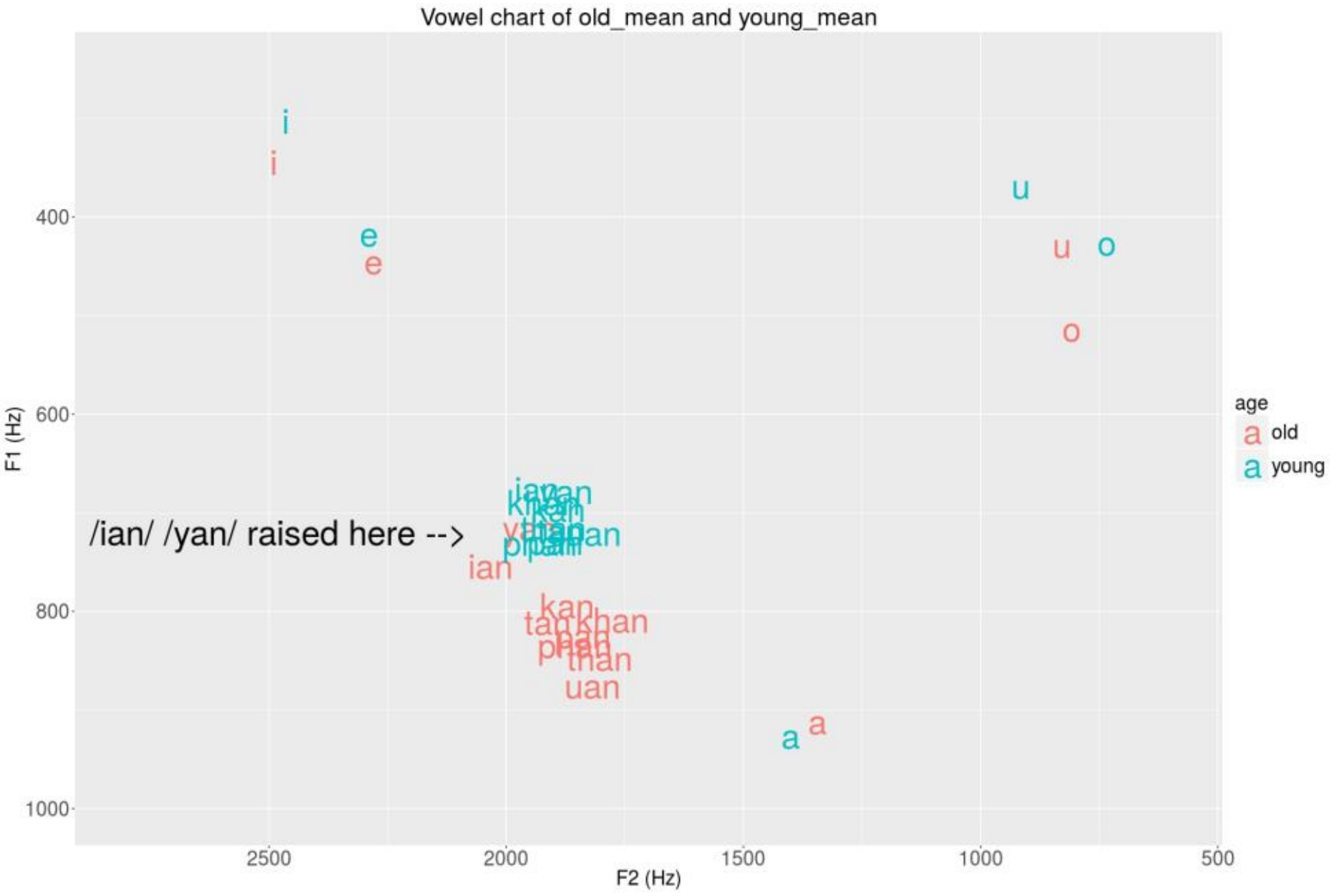

Figure 6: Combined vowel chart of the mean of two age groups.

3. Raising is influenced by some, but not all phonological environment, supporting only part of our third hypothesis:

   - There is no significant difference between the three environments: Diphthong, Aspirated and Unaspirated. Specifically, Aspirated environment does not have higher vowel height, providing no evidence for the aspirated unaspirated distinction in He and Rao (2013).
   - In the Diphthong environment, the height of /a/ in /yan/ is significantly higher than /uan/, lending support to categorize /uan/ differently from /yan/ and /ian/.

## 4 Discussion

Our results confirmed the observation in previous literature (He & Rao, 2013) that young speakers tend to raise the vowel /a/ in /an/ and young *female* speakers are leading the change. This is in accordance with Labov's statement that women are usually the innovators in unconscious sound change (Labov, 1990, pp. 215-218). However, there are several interesting points that deserve our attention.

### 4.1 Individual variances in raising

First, old female speaker No.1, who is younger than her male counterparts in the old age group (49 yrs vs. ca 57 yrs), seems to show unexpected raising (see Figure 7). This indicates that the vowel change may not be an absolute phenomenon, though it is manifested mostly in young speakers. It may also be found in some speakers from older generation.

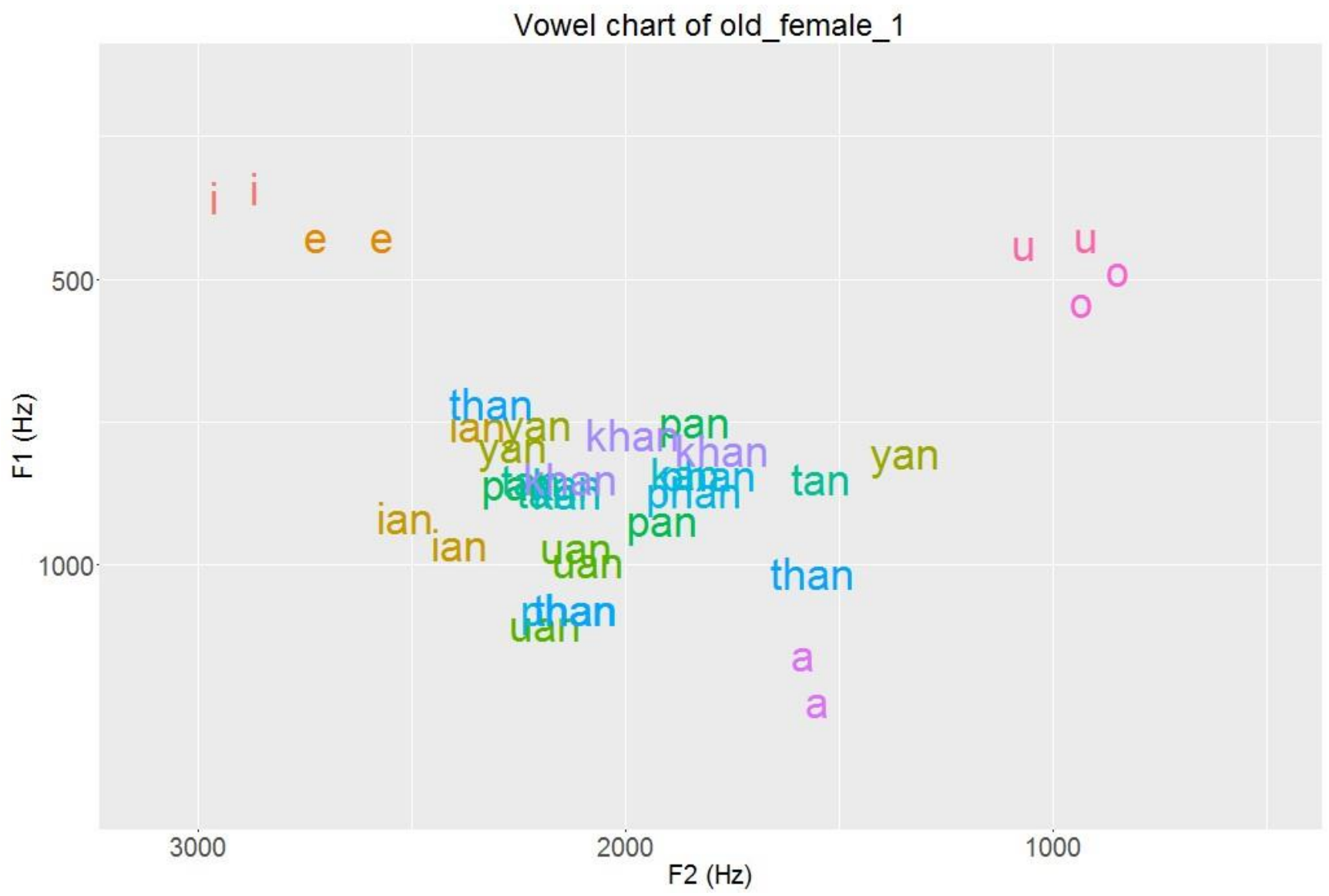

Figure 7: Some raising in old female speaker No.1.

The analysis of He and Rao (2013) shows that female speakers born after the 1980s have almost all raised /an/, whereas for male speakers only those born after the 1990s demonstrate the same raising effect. Our young male speakers are mostly born around 1990 so they can be said to be somewhat in between the "raising" and the "non-raising". Their data (as shown in Figure 3) is still significantly different from male speakers in the old age group, suggesting that most of them have raised /an/. The next step is to collect data from both middle-aged speakers and teenage speakers to determine where exactly the cut-off year may be.

At the same time, not all young speakers exhibit raising. Young female speaker No.1 for instance does not seem to raise /an/ at all (see Figure 8). This is much unexpected since young female speakers as innovators are more likely to show raising. It is more surprising considering the fact that this speaker has lived in near-downtown Chengdu before the age of 19 where standard Chengdu dialect is spoken and where He and Rao (2013) did their study. Her parents are also born and raised in Chengdu. So she should be very representative of the vowel raising effect. But our data suggests otherwise. One possible reason might be that the speaker went to college in Beijing and lived there for altogether 7 years.

## 4.2 Possible path of /a/ raising

As shown in Figure 4, the vowel height in /uan/ is different from /ian/ and /yan/. That is, although /i/, /y/ and /u/ are all high vowels, the [front] feature actually influences the height of /a/ considerably (t-test shows that in the old age group, the difference in F1 between /yan/ and /uan/ is significant).

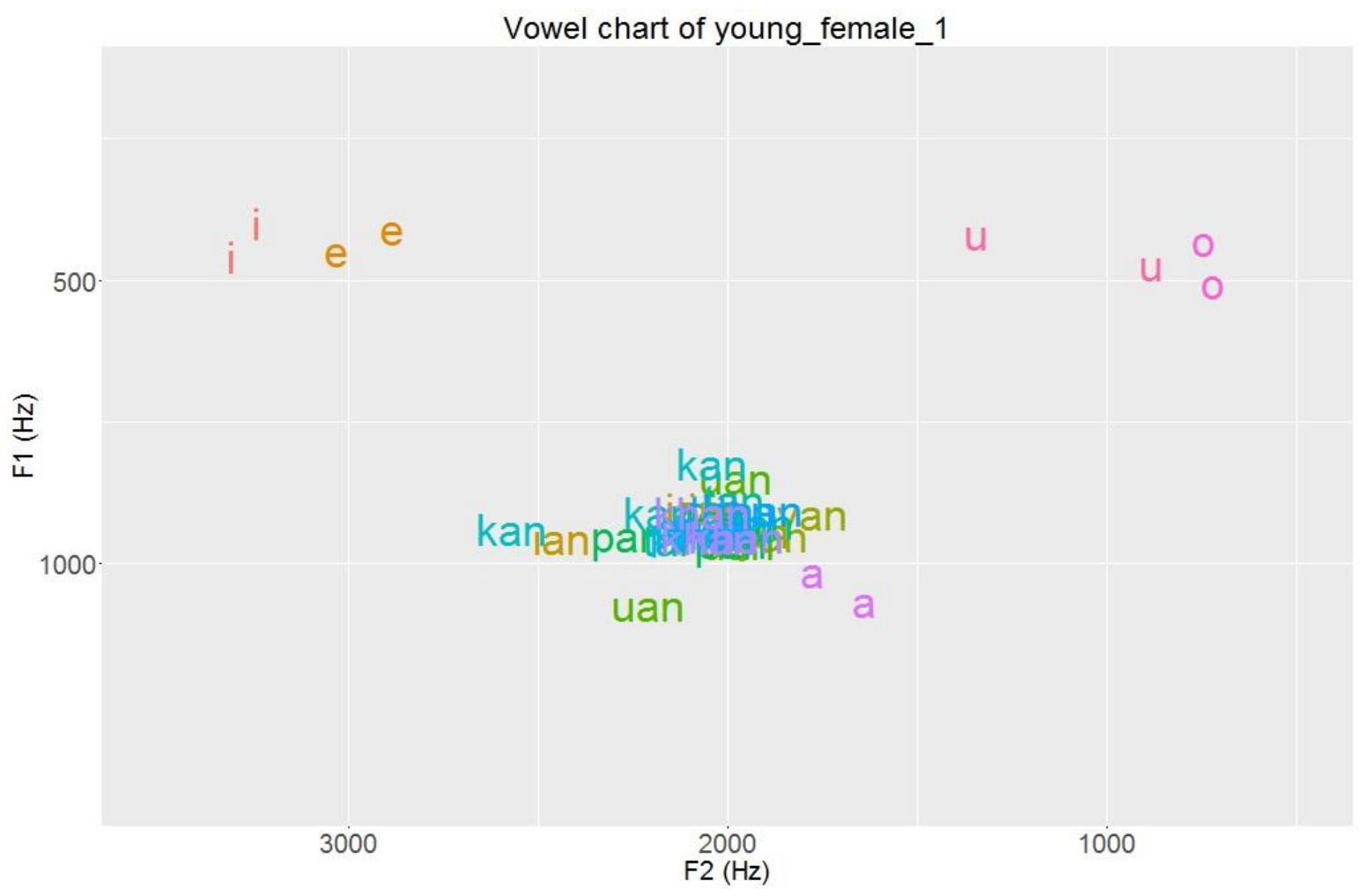

Figure 8: Unexpected: no raising for young female speaker No.1.

This result explains why in the previous literature summarized in Table 1, /uan/ never patterns with /ian/ or /yan/. In fact, Table 1 shows that /uan/ has long been categorized with /Can/. This is exactly the case for the old male speakers in this study, as shown in a more detailed vowel height plot of all the tokens (Figure 9).

Interestingly, Figure 6 suggests within the /Van/ condition, older speakers do raise /a/ in /ian/ and /yan/ (**ian**, **yan** is closer to **Xan**). We interpret this as /a/ first assimilates to its preceding front high vowels /i/ and /y/ for older speakers, which then becomes phonologized in younger speakers in all conditions, including /C$^h$an/ and /Can/. This shows a possible trajectory of the ongoing sound change in Chengdu dialect.

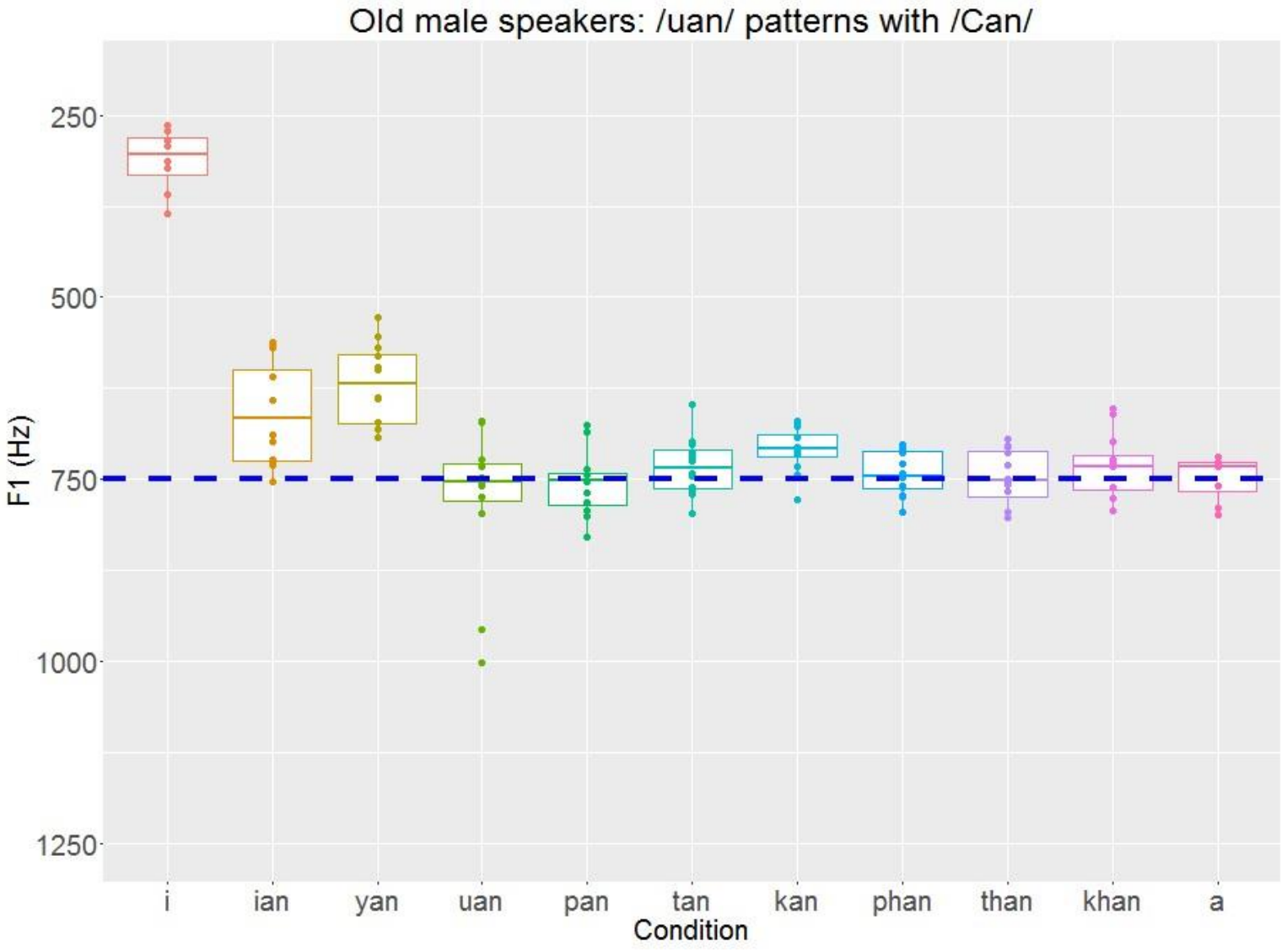

Figure 9: All tokens for old male speakers: /uan/ patterns with /Can/.

### 4.3 No distinction between aspirated & unaspirated

As suggested in He and Rao (2013), the aspirated environment is the first to be raised, according to their data obtained from speakers of 5 age groups. But the data in our study does not show any difference between aspirated and unaspirated environments, as shown in Figure 3 where the unaspirated (pink) and aspirated (yellow) are very close. If we zoom in and only look at these two environments (Figure 10 and 11), it is clearer that there is no systematic pattern between aspirated and unaspirated environments. That is, the F1 of both aspirated and unaspirated environments for old age group is right below the 750Hz line, whereas the F1 of both aspirated and unaspirated in young age group is just above the 750Hz line.

The only noticeable pattern seems to be the co-articulation effect of labial /p/, alveolar /t/ and velar /k/. That is, the height of /a/ seems to have the following order: /pan/ < /tan/ < /kan/ where < indicates ‘lower than’. We still need to further explore the reasons for this phenomena.

## 5 Conclusion and future work

To conclude, the raising of /a/ in young speakers of Chengdu dialect in /Van/, /Can/ and /$C^h$an/ are all attested. It is also clear from our data that female speakers lead the change. However, there is no difference between aspirated and unaspirated conditions.

Another finding is that the /a/ in /ian/ and /yan/ possible undergo raising first because of vowel height assimilation, which then spreads to other conditions such as /Can/. This is a potential path for the vowel raising phenomenon in Chengdu dialect.

In the future, we first need to obtain better quality recordings using more professional recording devices. Second, to understand the effect of age on vowel raising in greater detail, we need subjects from a more diverse range of ages. Third, to test whether there is a path for vowel raising, we need to analyze data from more age groups and see if there is a pattern that shows a clear path for raising.

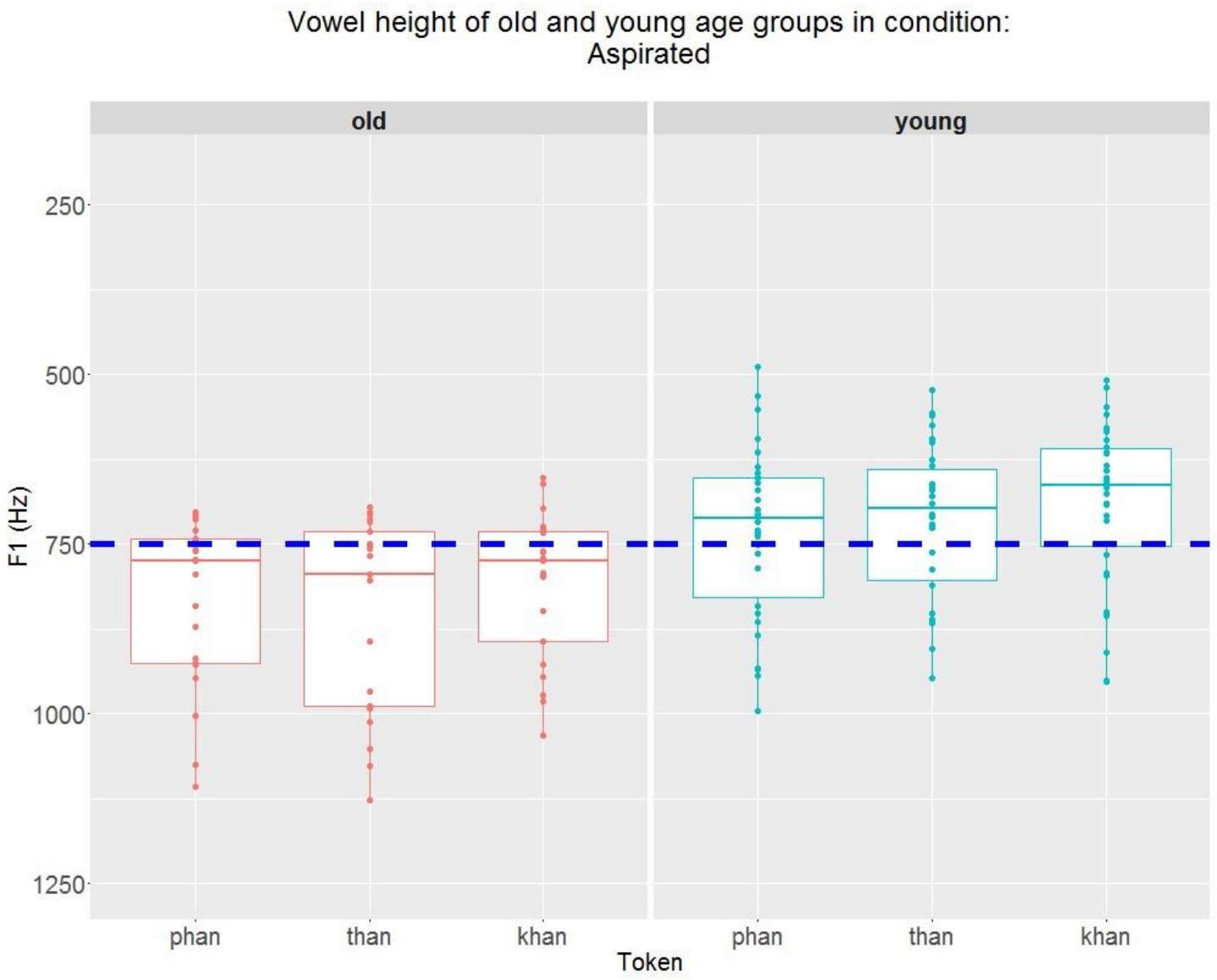

Figure 10: Aspirated environment.

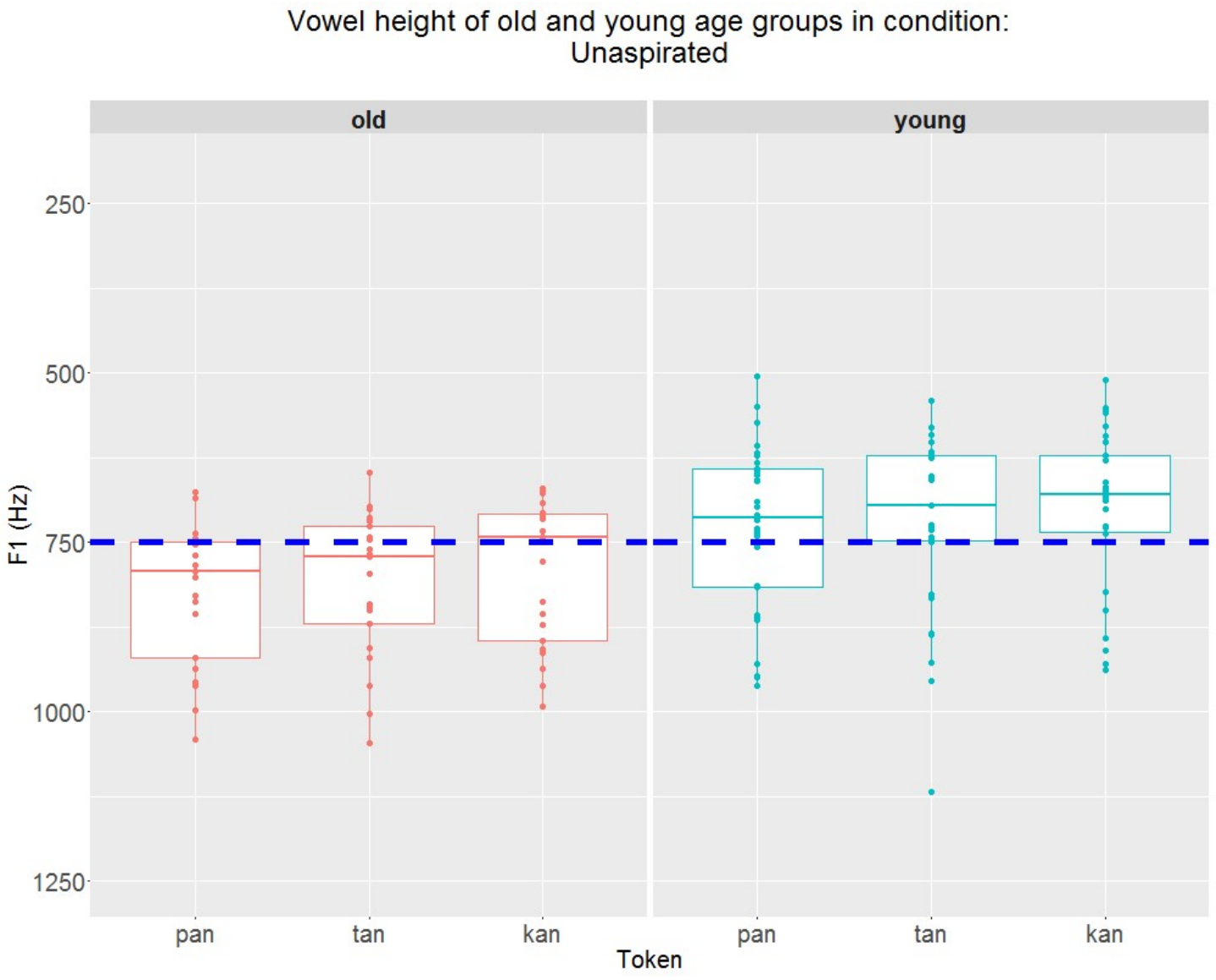

Figure 11: Unaspirated environment.